\pdfoutput=1 
\documentclass[letterpaper]{article} 
\usepackage[draft]{aaai25}  
\usepackage{times}  
\usepackage{helvet}  
\usepackage{courier}  
\usepackage[hyphens]{url}  
\usepackage{graphicx} 
\usepackage{natbib}  
\usepackage{caption} 
\usepackage{algorithm}
\usepackage{algorithmic}
\usepackage{amssymb}
\usepackage{amsmath}

\usepackage{newfloat}
\usepackage{cleveref}
\usepackage{listings}
\usepackage{caption}
\usepackage{subcaption}
\usepackage{multirow}
\DeclareCaptionStyle{ruled}{labelfont=normalfont,labelsep=colon,strut=off} 
\floatstyle{ruled}
\newfloat{listing}{tb}{lst}{}
\floatname{listing}{Listing}
\title{CLSP: High-Fidelity Contrastive Language-State Pre-training \\for Agent State Representation}

\nocopyright
\author{
    Fuxian Huang\equalcontrib,
    Qi Zhang\equalcontrib,
    Shaopeng Zhai\footnote{Co-corresponding authors},
    Jie Wang,
    Tianyi Zhang,
    \\Haoran Zhang,
    Ming Zhou, 
    Yu Liu, 
    Yu Qiao\footnotemark[2]
}
\affiliations {
    Shanghai AI Laboratory\\
    \{huangfuxian, zhangqi1, zhaishaopeng, wangjie, zhangtianyi, zhanghaoran, zhouming, liuyu, qiaoyu\}@pjlab.org.cn
}

\usepackage{bibentry}

\begin{document}

\maketitle
 
\begin{abstract}
With the rapid development of artificial intelligence, multimodal learning has become an important research area. For intelligent agents, the state is a crucial modality to convey precise information alongside common modalities like images, videos, and language. This becomes especially clear with the broad adoption of reinforcement learning and multimodal large language models. Nevertheless, the representation of state modality still lags in development. To this end,  we propose a High-Fidelity Contrastive Language-State Pre-training (CLSP) method, which can accurately encode state information into general representations for both reinforcement learning and multimodal large language models. Specifically, we first design a pre-training task based on the classification to train an encoder with coarse-grained information. Next, we construct data pairs of states and language descriptions, utilizing the pre-trained encoder to initialize the CLSP encoder. Then, we deploy contrastive learning to train the CLSP encoder to effectively represent precise state information. Additionally, we enhance the representation of numerical information using the Random Fourier Features (RFF) method for high-fidelity mapping. Extensive experiments demonstrate the superior precision and generalization capabilities of our representation, achieving outstanding results in text-state retrieval, reinforcement learning navigation tasks, and multimodal large language model understanding.
\end{abstract}

%

\section{Introduction}

Recently, multimodal data understanding, as the key of building AIGC and embodied AI,  has attracted increasing attention~\cite{zhu2023languagebind,kim2024openvla,chen2024far,zhan2024anygpt}.
As a crucial technology for understanding multimodal data, modality representation encodes data into a latent vector space to express information.
Current research on multimodal data mainly focuses on language, image, video, and audio data~\cite{zhu2023languagebind,zhu2024vision+,xie2024large}. However, in complex embodied AI environments, an agent's state often contains rich information that is vital for decision making, making it crucial to accurately represent and  understand the state.

Generally, the state of the embodied agent can be categorized as either vision-based or scalar-based. For vision-based state, the Contrastive Language–Image Pre-training (CLIP) \cite{radford2021learning}, a milestone representation method, can successfully estimate the similarity between a given image state and a text description. 
CLIP encodes image for both embodied AI and AIGC~\cite{fan2022minedojo, rocamonde2023vision, dang2023clip,liu2024improved} and has been widely studied and applied in various topics, such as MedCLIP\cite{wang2022medclip}, GeoCLIP\cite{cepeda2023geoclip}, CLIP-Art\cite{Conde_2021_CVPR}, MINECLIP\cite{fan2022minedojo}, etc. 
However, vision-based state only contains visually observable information, while scalar-based state information tends to encompass more complex and diverse numerical data, which is usually more difficult to perceive. 
Moreover, there has been insufficient research on the representation of scalar-based states. 
Additionally, we find that directly aligning states with text descriptions in a CLIP framework results in a loss of precision about the complex state information, especially when the prompt text is long (e.g., over 500 tokens) and contains complex information. This makes it difficult to retain accurate numerical information and its corresponding meanings. 
To successfully apply the representation to both multimodal large language models (LLMs) and reinforcement learning (RL), it needs to have both semantic understanding capabilities and informational precision. 
Therefore, two key issues need to be addressed: 1) extracting useful representations from complex scalar values is difficult; 2) the precision of learned representations corresponding to raw scalars tends to be low, hindering usability in subsequent tasks.

To tackle the above issues, in this work, we propose a novel method, High-Fidelity Contrastive Language-State Pre-training (CLSP) for Agent State Representation, which can accurately encode complex state information into general representations for both RL and multimodal LLMs. 
Specifically, CLSP represents scalars of various categories efficiently and accurately in two steps. 
In the first step, a supervised multiclass classification method is adopted to pre-train the encoder, which performs a coarse-grained division and classification prediction of scalar information across different categories.
The learned state encoder obtains coarse-grained information from the state and can provide a good initialization for further state-text alignment.
In the second step, similar to CLIP, we leverage contrastive learning\cite{khosla2020supervised}  to learn the alignment between CLSP encoder and text encoder. The CLSP encoder is initialized from the model learned in the first step and the text encoder is initialized from a pre-trained Bert model. 
The text data in state-text pairs contains rich and precise descriptive information, enabling state representations to have cross-modal capabilities and learn more fine-grained information.
Thus, the first issue can be effectively mitigated.
Moreover, to alleviate the second issue, we enhance the precision in learned representation with respect to the state scalars by utilizing a Random Fourier Features (RFF) strategy to encode scalars. 
In addition, we added MLPs after RFF to further improve the quality of encoded features.
In this way, the CLSP encoder is well-recognized and can encode high-fidelity scalar information in agent states. 
Finally, we conduct comprehensive experiments 
and verify the effectiveness of the representation via a series of retrieval tasks. 
To further validate the universality and accuracy of the representation, we conducted experimental verification on two tasks: RL navigation and multimodal LLMs tasks.
In the RL navigation task, we encode the state of the goal with the CLSP encoder and then use it as the target goal in the goal-conditioned RL. In the multimodal LLM task, we unify CLSP encoder and a pre-trained LLM for understanding the state.
Experimental results demonstrate the comprehensive performance of CLSP by accelerating the RL learning speed, increasing the final converged value, and reducing the error in scalar generation with multimodal LLM.

Our contributions can be summarized as follows:

\begin{itemize}
    \item We propose a novel framework, High-Fidelity Contrastive Language-State Pre-training, which consists of classification-based pre-training and contrastive learning-based state representation learning. To the best of our knowledge, this work pioneers the exploitation of agent state universal representation.
 
    \item To preserve the high-fidelity of important scalar values in the state, we deploy the random Fourier features method with MLPs  to encode the scalar values so as to improve the embedding precision.   
    \item We conducted comprehensive experiments on a FPS benchmark, which contains around 0.55 million collected state-text pairs. Moreover, we evaluate  CLSP encoder in two general downstream tasks. The overall experimental results validate the performance of our approach.
\end{itemize}

\section{Related Works}

\textbf{Contrastive Learning.}
Contrastive learning (CL) aims at learning meaningful representation by mapping similar samples closer in the feature space while pushing dissimilar ones apart.
There are two typical  CL methods: 1) augmentation-based method\cite{oord2018representation,chen2020simple,he2020momentum} learns the presentation by minimizing the distance between the positive sample and the corresponding augmented one while maximizing the distance with negative ones.
Because the information in the state is dense and interrelated, we did not conduct data augmentation for them.
2) CLIP-style method\cite{radford2021learning} samples a batch of \textit{image}-\textit{text} pairs and learns the representation by maximizing the similarity between corresponding \textit{image}-\textit{text} and minimizing the similarity between unpaired samples.
Our method is based on the second method, but learns the relationship between state and text. 

\begin{figure*}[!ht]
\centering
\includegraphics[width=0.95\textwidth]{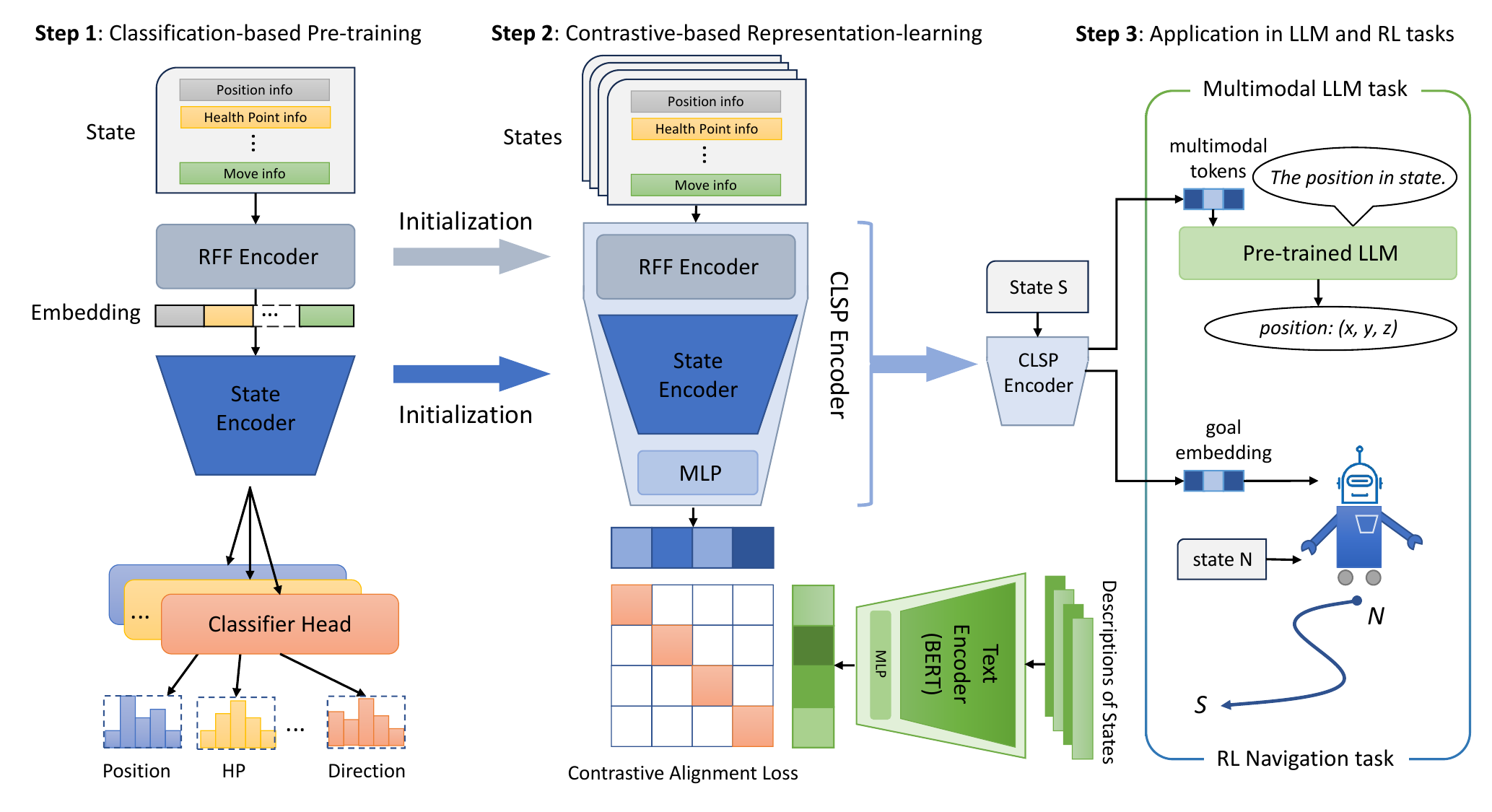} 
\vspace{-2mm}
\caption{The overall structure of CLSP has three steps. The first step is classification-based pre-training, which aims to obtain coarse-grained information. The second step is to align states and the corresponding text descriptions. The third step is to apply the learned CLSP to subsequent tasks.}
\label{fig:framework}
\vspace{-2mm}
\end{figure*}

\textbf{Multi-modal Language Model.} Recently, as research into large models continues to deepen, increasing attention has been paid to  multimodal language models. BLIP-2~\cite{li2023blip} uses a Q-former to connect a pre-trained image encoder with a large language model (LLM). Qwen-VL~\cite{bai2023qwen} employs cross-attention to link the pretrained image encoder and the LLM. LLava and the intern-VL series models~\cite{chen2024far,liu2024improved} achieve impressive multimodal understanding by connecting a pre-trained image encoder with an LLM using only a multilayer perceptron (MLP). Most current multimodal language models that perceive visual modalities utilize an image encoder pre-trained with CLIP, highlighting the effectiveness and importance of modality-specific pretrained encoders. LanguageBind~\cite{zhu2023languagebind} trains encoders for various modalities, and AnyGPT~\cite{zhan2024anygpt} simultaneously understands multiple modalities. However, the types of multimodal data currently include images, videos, audio, and language, while the state data in RL environments remains under-researched and primarily relies on textual descriptions, making it difficult to extend to long sequential records. As the application of multimodal language models in embodied intelligence gains increasing attention~\cite{kim2024openvla,fan2022minedojo,li2024llara}, the representation of state data is becoming increasingly important.

\textbf{Reinforcement Learning.} Generally, the state in RL falls into two types: 1) vision-based observation, where the environment includes Atari\cite{bellemare2013arcade}, Minecraft\cite{guss2019minerl}, ViZDoom\cite{Kempka2016ViZDoom},etc; 2) scalar-based observation, the environment includes MuJoCo\cite{todorov2012mujoco}, Dota 2\cite{berner2019dota}, OpenAI Gym Classic Control, etc.
For the first type, it is natural to deploy  CLIP to build the connection between image and text\cite{fan2022minedojo,rocamonde2023vision,dang2023clip}.
Specifically,\cite{fan2022minedojo} uses collected Minecraft videos and time-aligned transcripts to pretrain CLIP model, and then use it as the reward model to train the policy.
\cite{rocamonde2023vision} adopts a similar method but introduces a baseline regularization  to remove irrelevant parts of the CLIP representation.
\cite{dang2023clip,myers2023goal} utilizes CLIP to align the motion information presented in different observations with text. It is worth noting that although \cite{dang2023clip}  also uses scalar vectors to represent state, there are only a  few manually selected numbers, making it hard to generalize to different tasks. Our method focuses on  scenarios with only scalar-based observation.


\section{Methodology}

In this section, we present  CLSP to accurately encode state information into general representations for both RL and multimodal LLM.
As shown in \Cref{fig:framework}, the whole framework has three steps. Firstly,  the state encoder is pre-trained with a classification task. It is worth noting that the numbers in state is encoded with RFF to improve the numerical precision for further exploitation. 
Then the CLSP encoder and text encoder are trained based on contrastive learning, where the CLSP encoder is initialized with the parameters trained in   step 1.
Thirdly, the learned CLSP encoder can be applied in multimodal LLM task and RL navigation task. 

\subsection{Scalar  Encoded with Random Fourier Features }
Generally, the scalars in state are encoded by standard MLPs in RL. 
However, \citet{tancik2020fourier} shows that this encoding mechanism suffers from spectral bias, i.e., having difficulty  learning high-frequency details in scalar values.
It may result in a loss of information, especially for important scalars, such as \textit{position}, \textit{HP}, \textit{speed}, etc.
Therefore, we are seeking to employ an encoding method to project original scalars to a high-dimensional space, making it easier to capture complex information.
There are some common methods, such as Multi-Scale Normalizaer~\cite{springenberg2024offline}, NeRF positional encoding\cite{mildenhall2021nerf}, Random Fourier Feature (RFF)\cite{tancik2020fourier}, etc.
Multi-Scale Normalizer simply expands a scalar to a series of different fixed scales, while NeRF positional encoding encodes sequential data using sine and cosine functions. In contrast, Random Fourier Features (RFF) map scalars to a higher-dimensional space with randomly sampled Gaussian vectors, representing features across all selected possible frequencies. Therefore, we deploy RFF to extract useful information from the state, focusing on high-frequency components, specifically the details of scalars.
For a scalar value $v$, it is encoded in the way shown in \Cref{eq:rff}, where $\mathbf{b} \in \mathbb{R}^{d}$ is sampled from $\mathcal{N}\left(0, \sigma^2\right)$, and $\sigma$ is set to be 1 for simplicity and $d$ is chosen from $(4,6,8)$ according to the complexity of the scalar. 
For example, we choose $d=8$ for  \textit{position} because it varies continually in a large range (0,4000m), and $d=4$ for   \textit{HP} because it can only be  10 discrete values.
In addition,   we add two-layer MLPs after RFF to further improve the expressiveness of state.

\begin{equation}
\gamma(v)=[\cos (2 \pi \mathbf{b} v), \sin (2 \pi \mathbf{b} v)]^{\mathrm{T}}
\label{eq:rff}
\end{equation}

These encoded scalars provide a more complete presentation with more details for subsequent classification-based pre-training and contrastive-based representation learning.

\subsection{Classification-based Pre-training}

To capture the underlying patterns and semantic knowledge in states, we design a  classification-based pre-training task.
This task aims to predict some key properties in state using an approach similar to probing classifier\cite{belinkov2022probing}.
We select dozens of items from the state to predict their classes, including \textit{alive state}, \textit{position}, \textit{speed}, \textit{HP}, etc. 
These items naturally fall into two types: 1) \textbf{discrete values}, for example, \textit{alive state} has only three types: normal, down and dead; 2) \textbf{continuous values}, for example, \textit{position} represents the coordinates of any valid position. 
In practice, we discretize  these continuous values to maintain a relatively uniform distribution for the classification task.
In this way, we manually construct multiple classification labels $l$ for state $s$ concerning different items.
More details about the label construction are provided in the supplementary materials.
We employ a cross-entropy loss function to train this classifier, $L_{Pre} = -\sum_{c \in classes} l_c \log \left(p_c\right)
\label{eq:cross_entropy}$, where $p_c$ is the prediction of true class label $l_c$.
After pre-training on the collected data, the model gained a basic understanding of the state and can be fine-tuned for downstream tasks.

\subsection{Contrastive-based Representation learning}
In order to further improve the representation capability of state encoder, we adopt a contrastive-based learning method to align the state and text descriptions. 
Specifically, we jointly train a state encoder and a text encoder to learn the correlation between scalar-based states and text descriptions, where the state encoder is initialized from the pre-trained model introduced above and text encoder is initialized from DistilBERT\cite{sanh2019distilbert}.
During the training, we sample a batch of 
state-text pairs $\{s_i,t_i\}_{i=1}^B$, where $s_i$ represents the state and $t_i$ represents the corresponding text description.
The contrastive loss consists of two log classification terms, i.e., $state \rightarrow text$ loss $L_{S2T}$ and $text \rightarrow state$ loss $L_{T2S}$, which are defined  as follows:

\begin{equation}
L_{S 2 T}=-\frac{1}{B} \sum_{i=1}^B \log \frac{\exp \left(\phi (s_i^{\top}) \psi(t_i) / \tau\right)}{\sum_{j=1}^B \exp \left(\phi(s_i^{\top}) \psi(t_j) / \tau\right)},
\label{eq:loss_ls2t}
\end{equation}
\begin{equation}
L_{T 2 S}=-\frac{1}{B} \sum_{i=1}^B \log \frac{\exp \left(\psi(t_i^{\top}) \phi(s_i) / \tau\right)}{\sum_{j=1}^B \exp \left(\psi(t_i^{\top}) \phi(s_j) / \tau\right)},
\label{eq:loss_lt2s}
\end{equation}
where $\phi(s_i)$ is the encoded state by state encoder and $\psi(t_i)$ is the encoded text by text encoder, $\tau$ is a temperature hyperparameter. Additionally, we provide an algorithm table for CLSP in the supplementary materials.



\begin{figure}[t]
\centering
\includegraphics[width=0.95\columnwidth]{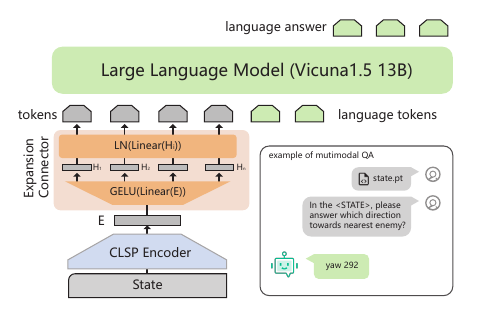} 
\caption{The architecture of multimodal LLM with our state encoder, and the example of state-based QA.}
\label{fig:state_llm}
\vspace{-4mm}
\end{figure}

\begin{table*}[!th]
\footnotesize
\begin{center}
\resizebox{1\hsize}{!}{
\begin{tabular}{l|ccc|cccc|ccc|ccc}
\hline
\multirow{2}{*}{Method}    & \multicolumn{1}{c|}{\multirow{2}{*}{R@1}} &  \multicolumn{1}{c|}{\multirow{2}{*}{R@5}}&  \multicolumn{1}{c|}{\multirow{2}{*}{R@10}}& \multicolumn{4}{c|}{Top-1 MAE (My)  }   & \multicolumn{3}{c|}{Top-1 MAE (Enemy)  }   & \multicolumn{3}{c}{Top-1 MAE (Teammate)  }   \\ \cline{5-14}
    & \multicolumn{1}{c|}{ }             & \multicolumn{1}{c|}{}             &              & \multicolumn{1}{c|}{HP}            & \multicolumn{1}{c|}{Position}       & \multicolumn{1}{c|}{Direction}      & Speed         & \multicolumn{1}{c|}{HP}            & \multicolumn{1}{c|}{Position}       & Distance       & \multicolumn{1}{c|}{HP}            & \multicolumn{1}{c|}{Position}       & Distance       \\ \hline
CLIP-Baseline   & \multicolumn{1}{c|}{0.00}          & \multicolumn{1}{c|}{4e-4}          &  6e-4         & \multicolumn{1}{c|}{22.46}         & \multicolumn{1}{c|}{1073.80}        & \multicolumn{1}{c|}{88.22}          & 6.23          & \multicolumn{1}{c|}{28.91}         & \multicolumn{1}{c|}{1062.49}        & 130.07         & \multicolumn{1}{c|}{26.98}         & \multicolumn{1}{c|}{1077.95}        & 148.75         \\ \hline
CLSP-Baseline & \multicolumn{1}{c|}{0.56}          & \multicolumn{1}{c|}{0.78}          & 0.81          & \multicolumn{1}{c|}{7.54}          & \multicolumn{1}{c|}{167.27}         & \multicolumn{1}{c|}{21.56}          & 1.44          & \multicolumn{1}{c|}{9.89}          & \multicolumn{1}{c|}{181.26}         & 40.01          & \multicolumn{1}{c|}{4.34}          & \multicolumn{1}{c|}{158.24}         & 26.54          \\ \hline
CLSP-MSN        & \multicolumn{1}{c|}{0.74}          & \multicolumn{1}{c|}{0.93}          & 0.94          & \multicolumn{1}{c|}{5.48}          & \multicolumn{1}{c|}{99.79}          & \multicolumn{1}{c|}{11.94}          & 0.78          & \multicolumn{1}{c|}{7.83}          & \multicolumn{1}{c|}{110.96}         & 28.33          & \multicolumn{1}{c|}{2.61}          & \multicolumn{1}{c|}{\textbf{87.16}} & \textbf{14.77} \\ \hline
CLSP-NPE        & \multicolumn{1}{c|}{0.67}          & \multicolumn{1}{c|}{0.94}          & 0.97          & \multicolumn{1}{c|}{6.17}          & \multicolumn{1}{c|}{126.04}         & \multicolumn{1}{c|}{9.57}           & \textbf{0.75} & \multicolumn{1}{c|}{8.11}          & \multicolumn{1}{c|}{143.32}         & 33.67          & \multicolumn{1}{c|}{4.67}          & \multicolumn{1}{c|}{122.96}         & 18.01          \\ \hline
CLSP-RFF        & \multicolumn{1}{c|}{0.73}          & \multicolumn{1}{c|}{0.94}          & 0.96          & \multicolumn{1}{c|}{5.13}          & \multicolumn{1}{c|}{121.77}         & \multicolumn{1}{c|}{29.77}          & 2.13          & \multicolumn{1}{c|}{9.52}          & \multicolumn{1}{c|}{145.89}         & 40.17          & \multicolumn{1}{c|}{5.38}          & \multicolumn{1}{c|}{122.47}         & 20.74          \\ \hline
CLSP-RFFM       & \multicolumn{1}{c|}{\textbf{0.83}} & \multicolumn{1}{c|}{\textbf{0.99}} & \textbf{0.99} & \multicolumn{1}{c|}{\textbf{3.15}} & \multicolumn{1}{c|}{\textbf{76.31}} & \multicolumn{1}{c|}{\textbf{12.46}} & 0.78          & \multicolumn{1}{c|}{\textbf{6.44}} & \multicolumn{1}{c|}{\textbf{90.15}} & \textbf{26.47} & \multicolumn{1}{c|}{\textbf{2.09}} & \multicolumn{1}{c|}{87.44}          & 18.97          \\ \hline
\end{tabular}
}
\vspace{-2mm}
\caption{Performance comparison for R@K and Top-1 MAE on contrastive learning task.}
\label{tab:contrastive_learning}
\end{center}
\vspace{-4mm}
\end{table*}

\subsection{Multimodal alignment with LLM}
To verify the representation capability of our state encoder in multimodal models, we design a large multimodal model to understand state information based on our state encoder and a pre-trained LLM. As shown in Figure ~\ref{fig:state_llm}, we use the Vicuna 1.5 13B~\cite{liu2024improved} model as the base large language model and design an expansion connector to convert the embeddings output by the state encoder into multiple tokens that can be understood by the large language model. Specifically, let the state embedding be  $ E \in \mathbb{R}^d $, and the output of the expansion connector as $ \{T_i\}_{i=1}^n \in \mathbb{R}^h $. The expansion connector can be expressed as:
\begin{equation}
H_i = \text{GELU}(\text{Linear}_i(E)) \quad (i = 1, 2, \ldots, n),
\label{eq:llm1}
\end{equation}
\begin{equation}
T_i = \text{LN}(\text{Linear}(H_i)) \quad (i = 1, 2, \ldots, n),
\label{eq:llm2}
\end{equation}
where \( H_i \in \mathbb{R}^{h'} \) is the hidden representation expanded from \( E \) for each output token, and LN represents layer normalization. In our experimental setup, \( d = 4096 \), \( h' = 256 \), \( n = 48 \), and \( h \) matches the input dimension of the large language model. The tokens \( \{T_i\}_{i=1}^n \) will be input into the large language model along with other language tokens, resulting in a generative output. Through the generated language output, the multimodal model can convert the information from the input state into language expressions.

During the training process, we proceed in two steps. In the first step, we freeze all  parameters of the LLM and the state encoder, training only the parameters of the expansion connector. In the second step, we freeze all  parameters of the LLM and train the parameters of both the expansion connector and the state encoder to further align the token representations.


\subsection{ RL Navigation with CLSP-encoded Goal}
To evaluate the effectiveness of CLSP, we choose a typical  RL task, namely the navigation task.
Specifically, we train an agent with policy $\pi$ to maximize its cumulative reward,  $\mathbb{E}_{a_t \sim \pi\left(\cdot \mid s_t, g\right),\left(s_t, a_t\right) \sim \tau}\left[\sum_{t=0}^T \gamma^t r\left(s_t, a_t, g\right)\right]$, where $r\left(s_t, a_t, g\right)$ is the reward indicating whether the agent completes the goal $g$. 
The goal $g$ is the encoded state from the CLSP encoder which contains the sampled target.
We employ Proximal Policy Optimization (PPO)\cite{schulman2017proximal} as the RL algorithm.


\section{Experiment}
In this section, we evaluate the effectiveness of CLSP across multiple tasks, including retrieval  task, multimodal LLM task, and RL navigation task.  

\subsection{Dataset $\&$ Environment}
Our experiments are mainly based on a multiplayer first-person shooting game  environment. 
The player's goal is to ensure  the survival of their team and eliminate encountered enemies.
Data is collected by running the trained policy model on 10 CPU servers over many rounds, resulting in approximately 17 hours of data.
To maintain a balanced data distribution, we sample approximately 18 minutes of data, resulting in about 0.55 million $state-text$ pairs.
After that, we generate the text descriptions for the sampled states. 
Then we randomly sample 5\% of them as the test dataset and the others as the training dataset. 
More details about the dataset are illustrated in the supplementary materials.
Additionally, we can customize the game environment to verify the effectiveness of our method in the RL task.

\begin{figure*}[htp]
     \centering
     \begin{subfigure}[b]{0.3\textwidth}
         \centering
         \includegraphics[width=\textwidth]{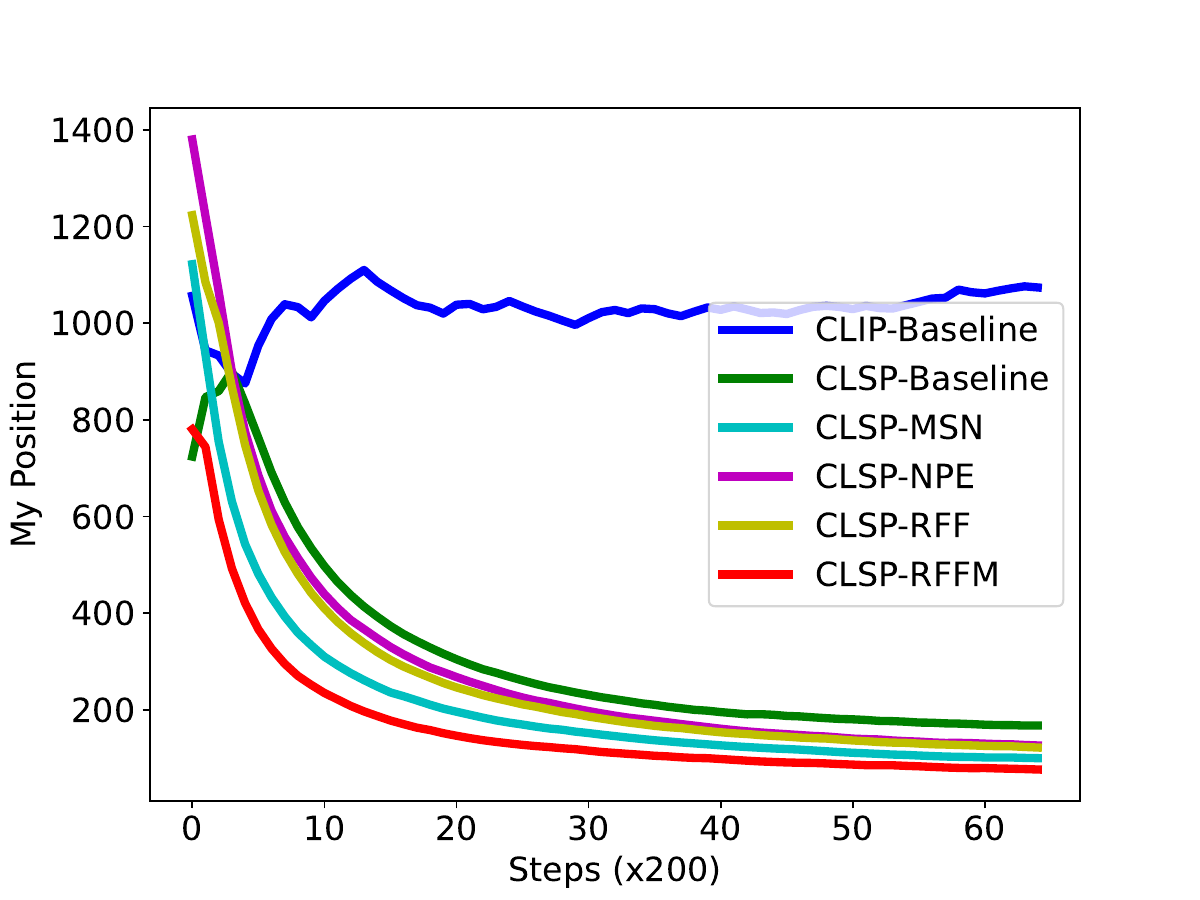}
         \caption{ My Position }
         \label{fig:retrieval_loss_1}
     \end{subfigure}
     \hspace{0.2pt}
     \begin{subfigure}[b]{0.295\textwidth}
         \centering
         \includegraphics[width=\textwidth]{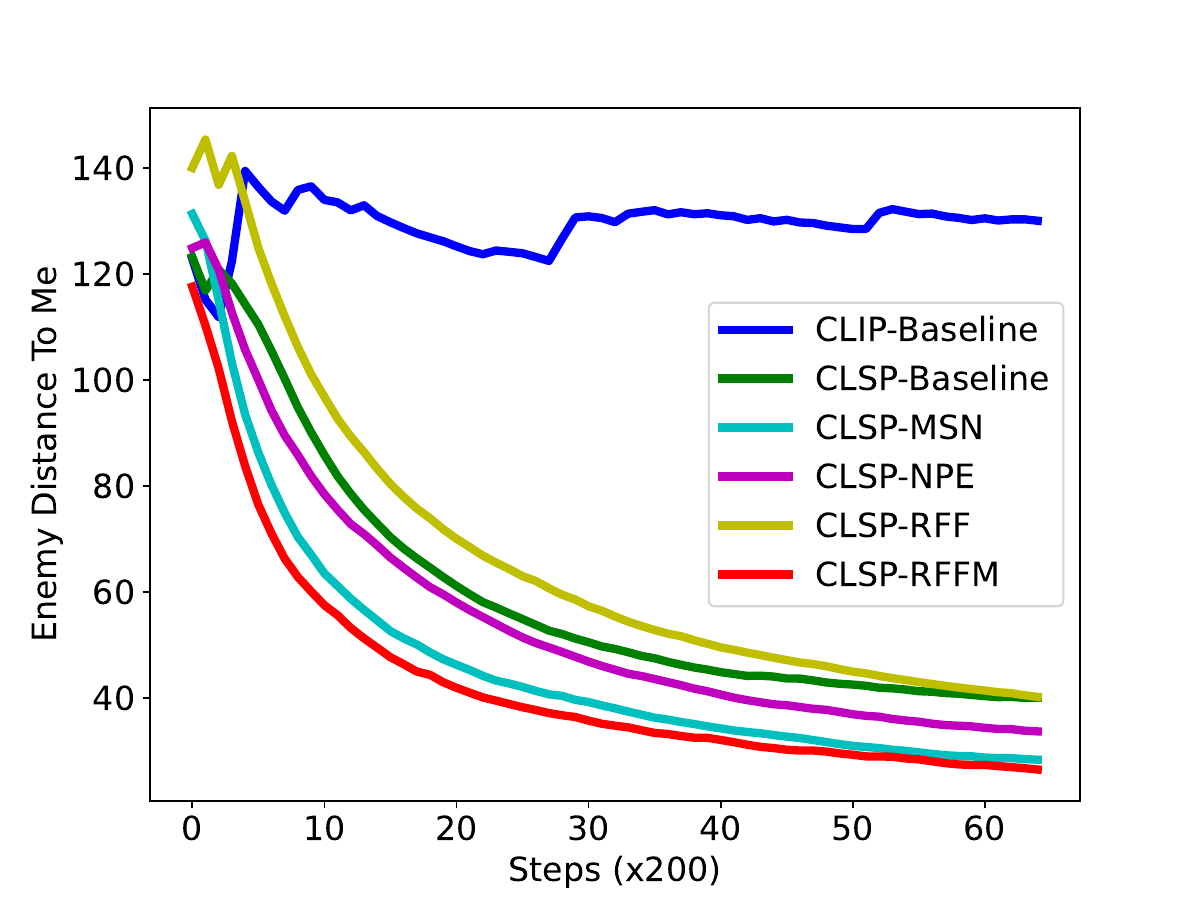}
         \caption{Enemy Distance}
         \label{fig:retrieval_loss_2}
     \end{subfigure}
     \hspace{0.2pt}
     \begin{subfigure}[b]{0.29\textwidth}
         \centering
         \includegraphics[width=\textwidth]{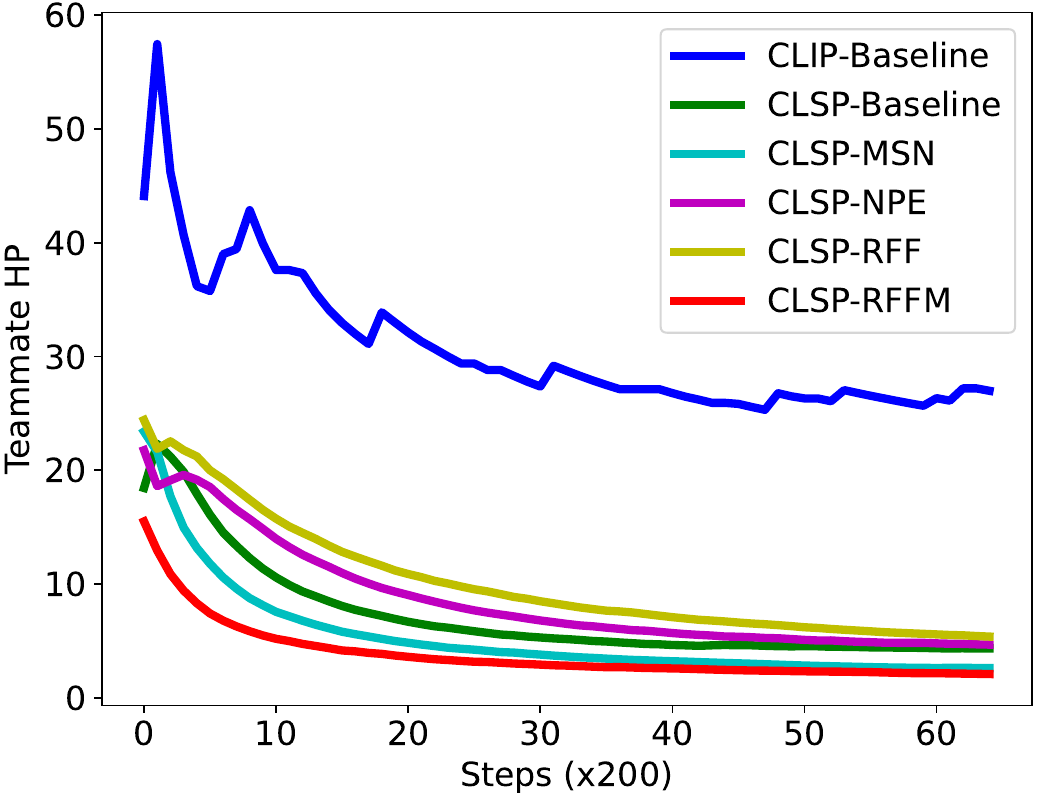}
         \caption{Teammate HP}
         \label{fig:retrieval_loss_3}
     \end{subfigure}
     \vspace{-2mm}
        \caption{Illustration of Top-1 MAE curves during the training procedure for different methods.}
        \label{fig:retrieval_loss}
        \vspace{-4mm}
\end{figure*}

\subsection{Metric}

\textbf{Top-1 MAE $\&$ R@K.} During the contrastive learning procedure, we periodically evaluate the performance of the model. 
First, we obtain a set of reference state representations generated by state encoder. 
Then, these states  representations are ranked according to their similarity to the query text description.
The Top-1  Mean Absolute Error (MAE), is defined as the mean difference between the top-1 matched state and the state corresponding to  the query text description.
Similarly, the R@K is  the mean ratio of successfully matched states in the top-K results based on the above similarity.
Lower Top-1 MAE and higher R@K indicate better performance.

\textbf{MedAE $\&$ MAE $\&$ RMSE.} In the multimodal model, the numerical values of various pieces of information generated for the state are compared with the annotated values to calculate the error, which serves as an evaluation metric for the accuracy of the generated information. Specifically, we use Median Absolute Error (MedAE), Mean Absolute Error (MAE), and Root Mean Square Error (RMSE)  to measure the generation error. Lower  values indicate better performance.

\textbf{ Goal completion ratio (GCR).} 
We further evaluate the effectiveness of our proposed approach in an RL navigation task.
In this task, the agent is randomly assigned a goal position at the start of each episode.
And it will receive a sparse reward when it achieves the goal, i.e., arriving at that position.
We consider the GCR as another indicator of learning performance. 
Higher GCR indicates better performance.

\begin{table*}[htp]
\footnotesize
\begin{tabular}{l|ccc|ccc|ccc|ccc}
\hline
\multicolumn{1}{c|}{\multirow{2}{*}{Method}} & \multicolumn{3}{c|}{HP}                                                         & \multicolumn{3}{c|}{Position}                                                              & \multicolumn{3}{c|}{Direction}                                                  & \multicolumn{3}{c}{Speed}                                                       \\ \cline{2-13} 
\multicolumn{1}{c|}{}                        & \multicolumn{1}{c|}{MedAE} & \multicolumn{1}{c|}{MAE}           & RMSE          & \multicolumn{1}{c|}{MedAE}          & \multicolumn{1}{c|}{MAE}            & RMSE           & \multicolumn{1}{c|}{MedAE} & \multicolumn{1}{c|}{MAE}           & RMSE          & \multicolumn{1}{c|}{MedAE} & \multicolumn{1}{c|}{MAE}           & RMSE          \\ \hline
w/o CLSP                                  & \multicolumn{1}{c|}{0}     & \multicolumn{1}{c|}{7.01}          & 18.86         & \multicolumn{1}{c|}{77.74}          & \multicolumn{1}{c|}{148.86}         & 268.56         & \multicolumn{1}{c|}{0}     & \multicolumn{1}{c|}{15.58}         & 29.47         & \multicolumn{1}{c|}{0}     & \multicolumn{1}{c|}{1.41}          & 2.01          \\ \hline
w/o CL-step                                      & \multicolumn{1}{c|}{0}     & \multicolumn{1}{c|}{2.42}          & 8.70          & \multicolumn{1}{c|}{25.32}          & \multicolumn{1}{c|}{65.26}          & 191.29         & \multicolumn{1}{c|}{0}     & \multicolumn{1}{c|}{3.95}          & 13.94         & \multicolumn{1}{c|}{0}     & \multicolumn{1}{c|}{0.53}          & 1.41          \\ \hline
w/o RFF                                       & \multicolumn{1}{c|}{0}     & \multicolumn{1}{c|}{2.08}          & 5.89          & \multicolumn{1}{c|}{24.68}          & \multicolumn{1}{c|}{49.13}          & 124.08         & \multicolumn{1}{c|}{0}     & \multicolumn{1}{c|}{2.96}          & 8.45          & \multicolumn{1}{c|}{0}     & \multicolumn{1}{c|}{0.27}          & \textbf{0.79} \\ \hline
CLSP-MLLM                                         & \multicolumn{1}{c|}{0}     & \multicolumn{1}{c|}{\textbf{0.78}} & \textbf{2.25} & \multicolumn{1}{c|}{\textbf{14.76}} & \multicolumn{1}{c|}{\textbf{18.67}} & \textbf{16.04} & \multicolumn{1}{c|}{0}     & \multicolumn{1}{c|}{\textbf{1.53}} & \textbf{5.65} & \multicolumn{1}{c|}{0}     & \multicolumn{1}{c|}{\textbf{0.19}} & 0.81          \\ \hline
\end{tabular}
\vspace{-2mm}
\caption{Performance of state understanding with multimodal LLM.}
\vspace{-6mm}
\label{tab:llm}
\end{table*}

\subsection{Baselines}
We  compare the contrastive learning performance among the following baselines. 
For fair comparison, we keep the text encoder in the same configuration and only study the influence of different initialization for the state encoder.
We test several different methods to encode the scalars in state, and they all share the same subsequent training procedure.
Thus, we named them as the variants of CLSP.

\textbf{CLIP-Baseline}: Directly using the CLIP framework, the images are replaced with states, and the parameters of state encoder are randomly initialized.

\textbf{CLSP-Baseline}: Only initializing state encoder with the model obtained from the classification-based pre-training stage without any method to encode the scalars in the state.

\textbf{CLSP-MSN}: Initialize state encoder with pre-trained classification model whose scalars are encoded by multi-scale normalizer (MSN)\cite{springenberg2024offline}

\textbf{CLSP-NPE}: Initialize state encoder with pre-trained classification model whose scalars are encoded by NeRF positional encoding (NPE)\cite{mildenhall2021nerf}.

\textbf{CLSP-RFF}: Initialize state encoder with pre-trained classification model whose scalars are encoded by vanilla random Fourier features (RFF) \cite{tancik2020fourier}.

\textbf{CLSP-RFFM}: Initialize state encoder with pre-trained classification model whose scalars are encoded RFF and learned MLPs.

\subsection{Implementation Details}
The state encoder consists of a fully connected layer followed by three Residual Blocks, and more details are in the supplementary materials.
We use pre-trained \textit{DistilBERT} as the text encoder. We use AdamW as the optimizer and set both the learning rate and weight decay to 1e-4.
For the contrastive objective, we set the temperature to $\tau=1$ and the batch size to 128.
In the RL navigation task, we use the PPO algorithm with Adam as the optimizer, $\gamma=0.995$, and a learning rate of 1e-4.
For the multimodal LLM task, we utilize Vicuna 1.5 13B~\cite{zheng2024judging} as the base LLM,  training Multimodal LLM with 534k state data, a batch size of 192, and a learning rate of 1e-3. 
We conduct the contrastive learning experiment  on 1 NVIDIA A100-SXM4-80GB GPU, multimodal LLM experiments on 6 NVIDIA A100-SXM4-80GB GPUs, and RL task on 10   AMD EPYC 7H12 64-Core CPU servers. 
More detailed descriptions of the experimental parameter settings are provided in the supplementary materials, and the code can be accessed\footnote{https://github.com/AnonymousUser360/CLSP}.

\subsection{Overall Results}

\begin{figure}[!t]
     \centering
         \centering
         \includegraphics[width=0.4\textwidth]{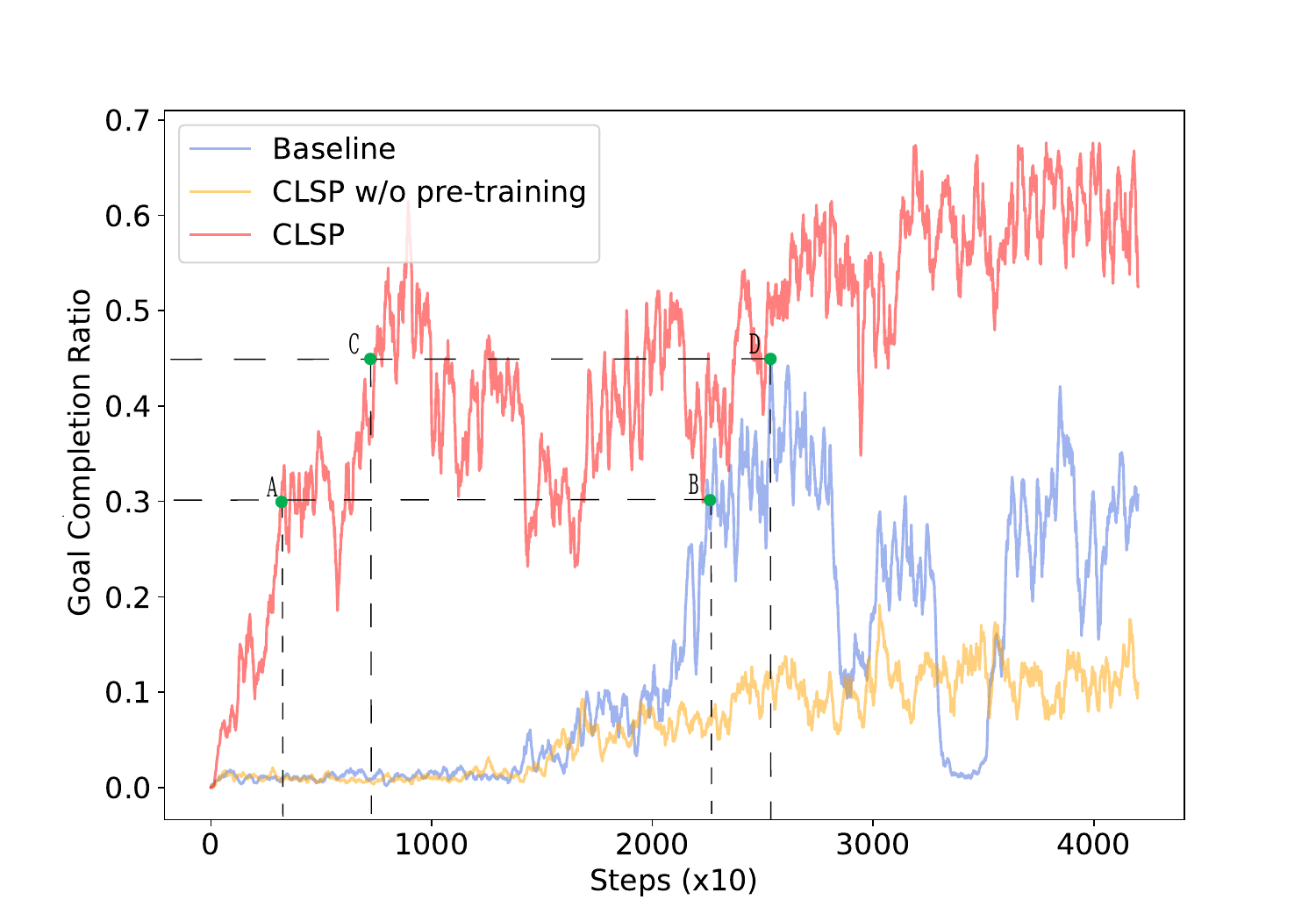}
         \vspace{-2mm}
         \caption{ Illustration of goal completion ratio during the training procedure for the RL navigation task. }
         \label{fig:rl_goal_ratio}

\vspace{-2mm}
\end{figure}
\begin{table}[!t]
\footnotesize
\begin{tabular}{l|cl|cl}
\hline
\multirow{2}{*}{Method} & \multicolumn{2}{c|}{GCR=0.3}                            & \multicolumn{2}{c}{GCR=0.45}                             \\ \cline{2-5} 
                        & \multicolumn{1}{c|}{RL}    & \multicolumn{1}{c|}{RL+CLSP}                    & \multicolumn{1}{c|}{RL}     &  \multicolumn{1}{c}{RL+CLSP}                    \\ \hline
CLSP                    & \multicolumn{1}{c|}{7.15M} & \multicolumn{1}{c|}{7.70M (A)} & \multicolumn{1}{c|}{18.43M} & \multicolumn{1}{c}{18.98M (C)} \\ \hline
Baseline                & \multicolumn{2}{c|}{63.25M (B)}                             & \multicolumn{2}{c}{68.75M (D)}                               \\ \hline
\end{tabular}
\caption{Comparison of data consumed for RL navigation task between CLSP and Baseline. RL+CLSP indicates the data includes the dataset to train CLSP. }
\label{tab:rl_navigation}
\vspace{-2mm}
\end{table}

\begin{table*}[t]
\footnotesize
\begin{tabular}{l|ccc|cccc|ccc|ccc}
\hline
    \multirow{2}{*}{\begin{tabular}[c]{@{}c@{}}Data\\ size\end{tabular}}                                                    & \multicolumn{1}{c|}{\multirow{2}{*}{R@1}} &  \multicolumn{1}{c|}{\multirow{2}{*}{R@5}}&  \multicolumn{1}{c|}{\multirow{2}{*}{R@10}}& \multicolumn{4}{c|}{Top-1 MAE (My)  }   & \multicolumn{3}{c|}{Top-1 MAE (Enemy)  }   & \multicolumn{3}{c}{Top-1 MAE (Teammate)  }                                                     \\ \cline{5-14}
 & \multicolumn{1}{c|}{ }             & \multicolumn{1}{c|}{ }             &              & \multicolumn{1}{c|}{HP}            & \multicolumn{1}{c|}{Position}        & \multicolumn{1}{c|}{Direction}      & Speed         & \multicolumn{1}{c|}{HP}             & \multicolumn{1}{c|}{Position}        & Distance       & \multicolumn{1}{c|}{HP}            & \multicolumn{1}{c|}{Position}        & Distance       \\ \hline
10\%       & \multicolumn{1}{c|}{2e-4}          & \multicolumn{1}{c|}{1e-3}         & 3e-3          & \multicolumn{1}{c|}{44.81}         & \multicolumn{1}{c|}{1126.90}          & \multicolumn{1}{c|}{91.85}          & 6.00          & \multicolumn{1}{c|}{35.94}          & \multicolumn{1}{c|}{1094.94}          & 118.01         & \multicolumn{1}{c|}{18.74}         & \multicolumn{1}{c|}{1129.46}          & 111.87          \\ \hline
25\%        & \multicolumn{1}{c|}{0.40}          & \multicolumn{1}{c|}{0.78}          & 0.89          & \multicolumn{1}{c|}{25.41} & \multicolumn{1}{c|}{602.68}          & \multicolumn{1}{c|}{72.77}          & 5.14          & \multicolumn{1}{c|}{25.68}          & \multicolumn{1}{c|}{616.81}          & 84.27          & \multicolumn{1}{c|}{12.34} & \multicolumn{1}{c|}{586.93}          & 78.73          \\ \hline
50\%      & \multicolumn{1}{c|}{0.45}          & \multicolumn{1}{c|}{0.83}          & 0.92          & \multicolumn{1}{c|}{18.16}         & \multicolumn{1}{c|}{344.13}          & \multicolumn{1}{c|}{61.81}          & 4.50          & \multicolumn{1}{c|}{17.41}          & \multicolumn{1}{c|}{357.64}          & 90.87          & \multicolumn{1}{c|}{7.36}          & \multicolumn{1}{c|}{324.34}          & 50.60          \\ \hline
100\%      & \multicolumn{1}{c|}{\textbf{0.66}} & \multicolumn{1}{c|}{\textbf{0.96}} & \textbf{0.99} & \multicolumn{1}{c|}{\textbf{8.75}}          & \multicolumn{1}{c|}{\textbf{199.53}} & \multicolumn{1}{c|}{\textbf{31.13}} & \textbf{2.18} & \multicolumn{1}{c|}{\textbf{10.92}} & \multicolumn{1}{c|}{\textbf{212.26}} & \textbf{48.80} & \multicolumn{1}{c|}{\textbf{5.84}}          & \multicolumn{1}{c|}{\textbf{198.24}} & \textbf{25.58}
 \\ \hline
\end{tabular}
\vspace{-2mm}
\caption{Ablation study of data size.}
\label{tab:ablation_data_size}
\vspace{-2mm}
\end{table*}

\begin{table*}[t]
\footnotesize
\begin{tabular}{l|ccc|cccc|ccc|ccc}
\hline
    \multirow{2}{*}{\begin{tabular}[c]{@{}c@{}}Batch\\ size\end{tabular}}                                                    & \multicolumn{1}{c|}{\multirow{2}{*}{R@1}} &  \multicolumn{1}{c|}{\multirow{2}{*}{R@5}}&  \multicolumn{1}{c|}{\multirow{2}{*}{R@10}}& \multicolumn{4}{c|}{Top-1 MAE (My)  }   & \multicolumn{3}{c|}{Top-1 MAE (Enemy)  }   & \multicolumn{3}{c}{Top-1 MAE (Teammate)  }                                                     \\ \cline{5-14}
 & \multicolumn{1}{c|}{ }             & \multicolumn{1}{c|}{ }             &              & \multicolumn{1}{c|}{HP}            & \multicolumn{1}{c|}{Position}        & \multicolumn{1}{c|}{Direction}      & Speed         & \multicolumn{1}{c|}{HP}             & \multicolumn{1}{c|}{Position}        & Distance       & \multicolumn{1}{c|}{HP}            & \multicolumn{1}{c|}{Position}        & Distance       \\ \hline
16                                                   & \multicolumn{1}{c|}{0.01}          & \multicolumn{1}{c|}{0.05}         & 0.10          & \multicolumn{1}{c|}{29.12}         & \multicolumn{1}{c|}{443.27}          & \multicolumn{1}{c|}{84.56}          & 6.05          & \multicolumn{1}{c|}{26.79}          & \multicolumn{1}{c|}{467.93}          & 112.44         & \multicolumn{1}{c|}{19.53}         & \multicolumn{1}{c|}{491.29}          & 31.18          \\ \hline
32                                                   & \multicolumn{1}{c|}{0.30}          & \multicolumn{1}{c|}{0.70}          & 0.84          & \multicolumn{1}{c|}{\textbf{7.59}} & \multicolumn{1}{c|}{444.50}          & \multicolumn{1}{c|}{54.47}          & 3.74          & \multicolumn{1}{c|}{17.59}          & \multicolumn{1}{c|}{456.09}          & 87.41          & \multicolumn{1}{c|}{\textbf{4.68}} & \multicolumn{1}{c|}{438.32}          & 32.81          \\ \hline
64                                                   & \multicolumn{1}{c|}{0.51}          & \multicolumn{1}{c|}{0.86}          & 0.94          & \multicolumn{1}{c|}{16.25}         & \multicolumn{1}{c|}{206.01}          & \multicolumn{1}{c|}{49.91}          & 3.46          & \multicolumn{1}{c|}{18.82}          & \multicolumn{1}{c|}{259.48}          & 64.55          & \multicolumn{1}{c|}{5.37}          & \multicolumn{1}{c|}{206.27}          & 26.85          \\ \hline
128                                                  & \multicolumn{1}{c|}{\textbf{0.66}} & \multicolumn{1}{c|}{\textbf{0.96}} & \textbf{0.99} & \multicolumn{1}{c|}{8.75}          & \multicolumn{1}{c|}{\textbf{199.53}} & \multicolumn{1}{c|}{\textbf{31.13}} & \textbf{2.18} & \multicolumn{1}{c|}{\textbf{10.92}} & \multicolumn{1}{c|}{\textbf{212.26}} & \textbf{48.80} & \multicolumn{1}{c|}{5.84}          & \multicolumn{1}{c|}{\textbf{198.24}} & \textbf{25.58}
 \\ \hline
\end{tabular}
\vspace{-2mm}
\caption{Ablation study of batch size.}
\label{tab:ablation_batch_size}
\vspace{-2mm}
\end{table*}

\begin{table*}[!h]
\footnotesize
\resizebox{1\hsize}{!}{
\begin{tabular}{l|ccc|cccc|ccc|ccc}
\hline
         \multirow{2}{*}{\begin{tabular}[c]{@{}c@{}}Classifier\\ target\end{tabular}}   & \multicolumn{1}{c|}{\multirow{2}{*}{R@1}} &  \multicolumn{1}{c|}{\multirow{2}{*}{R@5}}&  \multicolumn{1}{c|}{\multirow{2}{*}{R@10}}& \multicolumn{4}{c|}{Top-1 MAE (My)  }   & \multicolumn{3}{c|}{Top-1 MAE (Enemy)  }   & \multicolumn{3}{c}{Top-1 MAE (Teammate)  }                                                     \\ \cline{5-14}
 & \multicolumn{1}{c|}{ }             & \multicolumn{1}{c|}{ }             &              & \multicolumn{1}{c|}{HP}            & \multicolumn{1}{c|}{Position}        & \multicolumn{1}{c|}{Direction}      & Speed         & \multicolumn{1}{c|}{HP}             & \multicolumn{1}{c|}{Position}        & Distance       & \multicolumn{1}{c|}{HP}            & \multicolumn{1}{c|}{Position}        & Distance       \\ \hline
Self                                                      & \multicolumn{1}{c|}{0.33 }         & \multicolumn{1}{c|}{0.74 }         & 0.88         & \multicolumn{1}{c|}{9.67}         & \multicolumn{1}{c|}{326.38}           & \multicolumn{1}{c|}{33.12}          & \textbf{1.73}         & \multicolumn{1}{c|}{14.57}          & \multicolumn{1}{c|}{337.10}           & 88.59          & \multicolumn{1}{c|}{13.75}         & \multicolumn{1}{c|}{367.76}           & 90.79         \\ \hline
Team                                                      & \multicolumn{1}{c|}{0.45}          & \multicolumn{1}{c|}{0.81}          & 0.89          & \multicolumn{1}{c|}{12.85}         & \multicolumn{1}{c|}{\textbf{121.87}} & \multicolumn{1}{c|}{47.08}          & 3.62          & \multicolumn{1}{c|}{15.31}          & \multicolumn{1}{c|}{\textbf{167.59}} & 53.29          & \multicolumn{1}{c|}{6.28}          & \multicolumn{1}{c|}{\textbf{121.17}} & \textbf{22.86} \\ \hline
Enemy                                                     & \multicolumn{1}{c|}{0.49}          & \multicolumn{1}{c|}{0.87}          & 0.95          & \multicolumn{1}{c|}{11.27}         & \multicolumn{1}{c|}{272.11}          & \multicolumn{1}{c|}{38.85}          & 3.44          & \multicolumn{1}{c|}{16.32}          & \multicolumn{1}{c|}{285.90}          & \textbf{48.39} & \multicolumn{1}{c|}{14.25}         & \multicolumn{1}{c|}{322.39}          & 85.78          \\ \hline
All                                                       & \multicolumn{1}{c|}{\textbf{0.66}} & \multicolumn{1}{c|}{\textbf{0.96}} & \textbf{0.99} & \multicolumn{1}{c|}{\textbf{8.75}} & \multicolumn{1}{c|}{199.53}          & \multicolumn{1}{c|}{\textbf{31.13}} & 2.18 & \multicolumn{1}{c|}{\textbf{10.92}} & \multicolumn{1}{c|}{212.26}          & 48.80          & \multicolumn{1}{c|}{\textbf{5.84}} & \multicolumn{1}{c|}{198.24}          & 25.58  
 \\ \hline
\end{tabular}
}
\vspace{-2mm}
\caption{Ablation study of classifier target.}
\label{tab:ablation_classifier_target}
\vspace{-2mm}
\end{table*}

In our experiments, we  conduct state-text alignment training task and evaluate different methods based on R@K and Top-1 MAE. 
\Cref{tab:contrastive_learning} presents some quantitative  results and we can find that: 
1) Classification-based pre-training is critical for the state-text contrastive learning task. 
Specifically, compared to CLIP-Baseline, CLSP-Baseline improves the recall scores from nearly zero to 71.6\% on average and reduces the Top-1 MAE by 77.3\% on average.
These results demonstrate the effectiveness of coarse-grained information obtained by state encoder through classification-based  pre-training.
2) All the variants of CLSP  consistently outperform CLSP-Baseline by a large margin in  recall score and Top-1 MAE, which shows the importance of encoding the scalar values to maintain higher scalar precision.
For example, CLSP-MSN achieves a higher recall score by around 15\% on average than CLSP-Baseline, and decreases the Top-1 MAE by around 37.6\% on average. 
3) CLSP-RFFM achieves the best results in 10/13 items among  CLSP variants, demonstrating its effectiveness in adding learnable MLPs after a traditional RFF encoder.
Additionally, \Cref{fig:retrieval_loss} illustrates the training curves for three Top-1 MAE: My Position, Enemy Distance, and Teammate HP.
We can see that CLSP-Baseline and other  variants of CLSP can consistently reduce the Top-1 MAE significantly.
Especially, CLSP-RFFM can achieve lower loss at a faster rate, which proves the effectiveness of our proposed CLSP training framework.

\subsection{Results of Multimodal LLM} 

To further validate the representation performance of our CLSP encoder in multimodal LLMs, we perform experiments on several encoder variants. The overall results of these  encoders are presented in Table~\ref{tab:llm}. The results show that our CLSP encoder significantly outperforms other types of encoders while preserving the precision of information.
Specifically, the method labeled CLSP-MLLM indicates the integration of our CLSP encoder with the multimodal LLM, as depicted in Figure~\ref{fig:state_llm}. The variant ``w/o CLSP'' refers to using a state encoder learned directly from a randomly initialized model without the CLSP encoder during multimodal LLM training. ``w/o CL-step'' refers to the absence of training step 2, which involves contrastive  representation learning, utilizing only a classification-based pre-trained state encoder as the multimodal encoder. ``w/o RFF'' means CLSP without Random Fourier Features.
As shown in the results table, the CLSP encoder demonstrates high-fidelity capabilities. The Mean Absolute Error (MAE) for health information is reduced by approximately 62.5\% compared to the best baseline, and the MAE for position information is reduced by about 62.0\%. Additionally, the Median Absolute Error (MedAE) for health points, direction, and speed is zero, indicating that all methods  generate a large amount of accurate information. Meanwhile, the significantly reduced average error of CLSP further demonstrates the model's excellent generalization performance, enabling it to handle outliers and out-of-distribution (OOD) data effectively.

\subsection{Results of RL Navigation} 
In the RL navigation task, we  evaluate the effectiveness of the goal state embedding by comparing the GCR with baselines.
Higher GCR indicates better performance.
In this task, CLSP uses the embedding of the goal state produced by the trained state encoder as the goal.
CLSP w/o pre-training is the same as CLSP except that the parameters are initialized randomly.
Baseline denotes using the raw goal position, i.e., $(x,y,z)$, as the goal.
As shown in \Cref{fig:rl_goal_ratio}, CLSP achieves a much faster learning speed and converges to a higher GCR of around 60\%, which is significantly better than  CLSP w/o pre-training and Baseline.
These results demonstrate that the training pipeline and proposed goal representation are both beneficial for RL navigation task.
\Cref{tab:rl_navigation} shows the data costs for CLSP and Baseline to learn the navigation task. 
Since it takes around 0.55 million (M) states in advance to train CLSP,   we also list the data used  by RL learning and previous CLSP training for a fair comparison, namely RL+CLSP. 
Additionally, points A and B in \Cref{fig:rl_goal_ratio}  correspond to the results of GCR=0.3 in \Cref{tab:rl_navigation}. 
It can be seen that CLSP only requires around 12\% of the data to achieve  a similar result to Baseline.

\subsection{Ablation Study}
We conduct   ablation studies on different hyper-parameters, i.e. the data size, the batch size and the target of the classifier. 
These results are collected after training for one epoch.

\textbf{Effects of different data size.} In \Cref{tab:ablation_data_size}, we investigate the impact of different data sizes on the contrastive learning task.
The results demonstrate that more data   consistently improves the recall score and reduces the Top-1 MAE.
Specifically, when increasing the data size from $10\%$ to $25\%$ of the total, the R@10   significantly increases from 3e-1 to 0.89.

\textbf{Effects of different batch sizes.} \Cref{tab:ablation_batch_size}
 illustrates the impacts of different batch sizes on the contrastive learning task. 
 It can be observed that as the batch size increases, the recall scores consistently improve. 
 Additionally,  Top-1 MAE significantly decreases  as the batch size increases.
 
\textbf{Effects of different classification targets.} In the proposed method, classification-based pre-training plays an important role in improving the understanding of state by classifying the items in state.
We study the impacts of classifying different types of items in the state. 
As shown in \Cref{tab:ablation_classifier_target}, classifying partial targets  items can yield better results on Top-1 MAE. However, classifying all items significantly outperforms other methods on most metrics, especially for R@1, R@5 and R@10.

\begin{figure}[th] 
     \centering
     \includegraphics[width=0.49\textwidth]{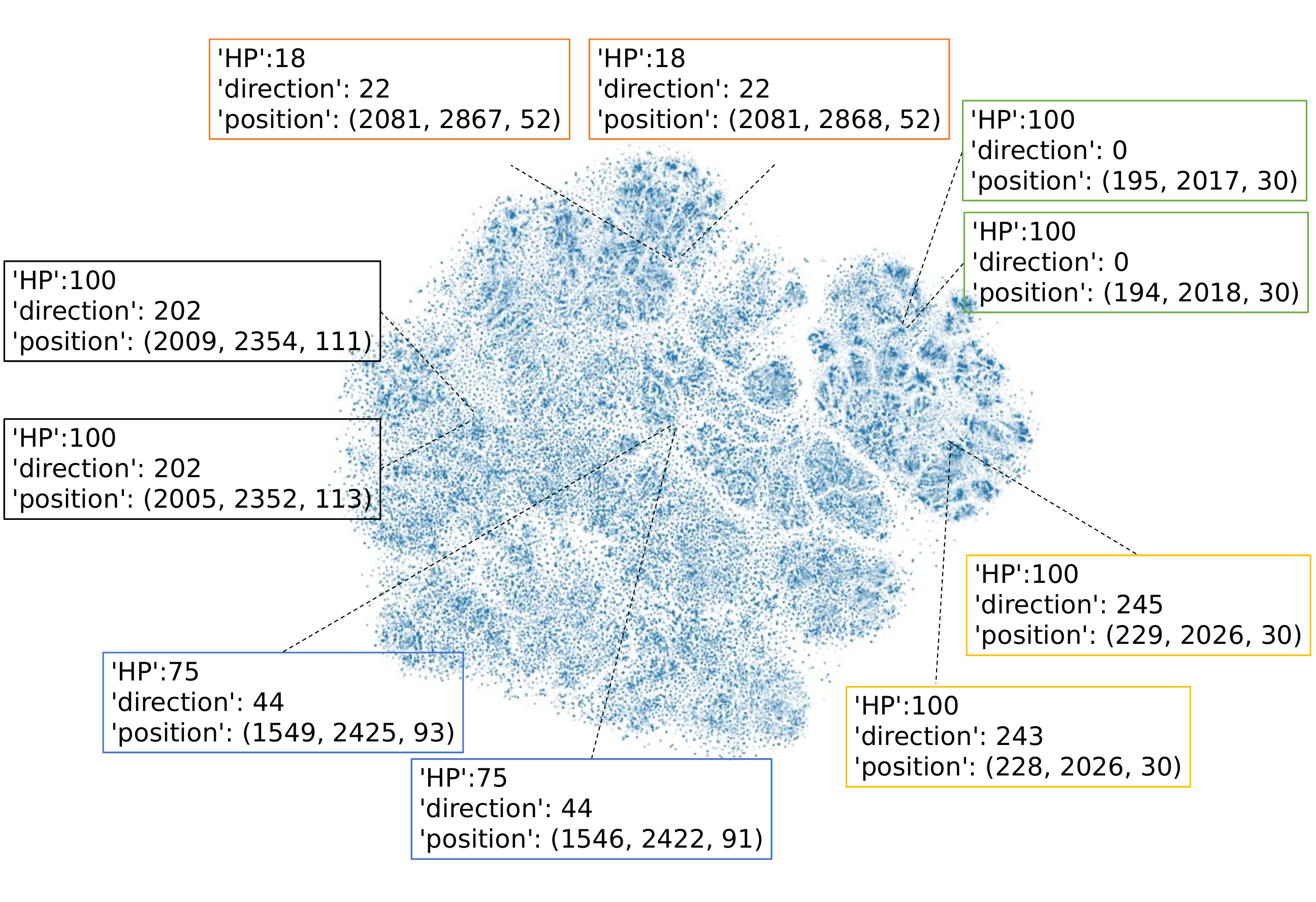}
     \vspace{-4mm}
     \caption{ The t-SNE visualization of state representations produced by CLSP. }
     \label{fig:tsne}
\vspace{-4mm}
\end{figure}

\subsection{Visualization of Distinguishable Representations}
\Cref{fig:tsne} illustrates the two-dimensional t-SNE embedding of the state representations produced by CLSP. 
We  select ten states and provide more information about them, where  similar states are close in the state space.
As shown in \Cref{fig:tsne}, the properties of adjacent points exhibit similar properties with respect to HP, direction or position.
This phenomenon demonstrates that the state embeddings produced by CLSP are smooth.

\section{Conclusion}
In this paper, we present CLSP framework to address the underdeveloped area of state modality representation in multimodal learning and RL. By designing a classification-based pre-training process and utilizing contrastive learning, our approach effectively encodes state information into general high-fidelity 
representations, enhancing the alignment between states and language descriptions. The use of RFF enhances the precision in representing numerical data even further. Extensive experiments demonstrate outstanding results in text-state retrieval, RL navigation, and multimodal understanding tasks, highlighting  CLSP's potential to  intelligent agents and multimodal LLMs. 
\bibliography{aaai25}

\end{document}